\documentclass[]{fairmeta}

\usepackage{amsmath}
\usepackage{amssymb}
\usepackage{algorithm}
\usepackage[noend]{algpseudocode}

\usepackage{xspace}
\usepackage{adjustbox}
\usepackage{enumitem}
\usepackage{soul}
\usepackage{url}

\newcommand{\custompara}[1]{{\vspace{1mm}\noindent\textbf{#1}\xspace}}

\makeatletter
\DeclareRobustCommand\onedot{\futurelet\@let@token\@onedot}
\def\@onedot{\ifx\@let@token.\else.\null\fi\xspace}
\def\eg{\emph{e.g}\onedot}
\def\ie{\emph{i.e}\onedot}

\makeatother

\title{Learning to Reason by Analogy via Retrieval-Augmented Reinforcement Fine-Tuning}

\author[1,2,*]{Zilin Xiao}
\author[1]{Qi Ma}
\author[1]{Jason Chen}
\author[1]{Xintao Chen}
\author[1]{Avinash Atreya}
\author[2]{Hanjie Chen}
\author[2]{Vicente Ordonez}

\affiliation[1]{Meta Superintelligence Labs}
\affiliation[2]{Rice University}

\contribution[*]{Work partially done during internship at Meta}

\abstract{
Retrieval-augmented generation (RAG) has become a standard mechanism for grounding language models in external knowledge, yet conventional retrieval based on lexical or semantic similarity is poorly suited for complex reasoning tasks: a semantically similar problem may demand an entirely different solution strategy, while a superficially different problem may share the same underlying reasoning pattern.
We propose \textbf{Retrieval-Augmented Reinforcement Fine-Tuning (RA-RFT)}, a post-training framework that teaches language models to reason by analogy.
RA-RFT uses gold-relevance distillation to train a retriever that ranks contexts by expected reasoning benefit rather than semantic overlap, and then fine-tunes the policy model via reinforcement fine-tuning methods with retrieved analogous demonstrations, so the model learns to leverage reasoning traces under verifiable outcome rewards.
We further analyze the diversity of retrieved contexts and find that reasoning-aware retrieval surfaces complementary solution strategies that provide distinct reasoning scaffolds for individual problems.
Across challenging mathematical reasoning benchmarks, RA-RFT consistently outperforms standard reinforcement fine-tuning methods. For example, it improves AIME 2025 average@32 accuracy by 7.1 and 2.8 points over GRPO for Qwen3-1.7B and Qwen3-4B respectively---suggesting that reasoning-aware retrieval is a complementary axis of improvement and orthogonal to advances in reward design or training curricula.
}

\date{\today}
\correspondence{Zilin Xiao at \email{zilin@meta.com}}

\begin{document}

\maketitle

\begin{figure*}[t]
\centering
\includegraphics[width=\textwidth]{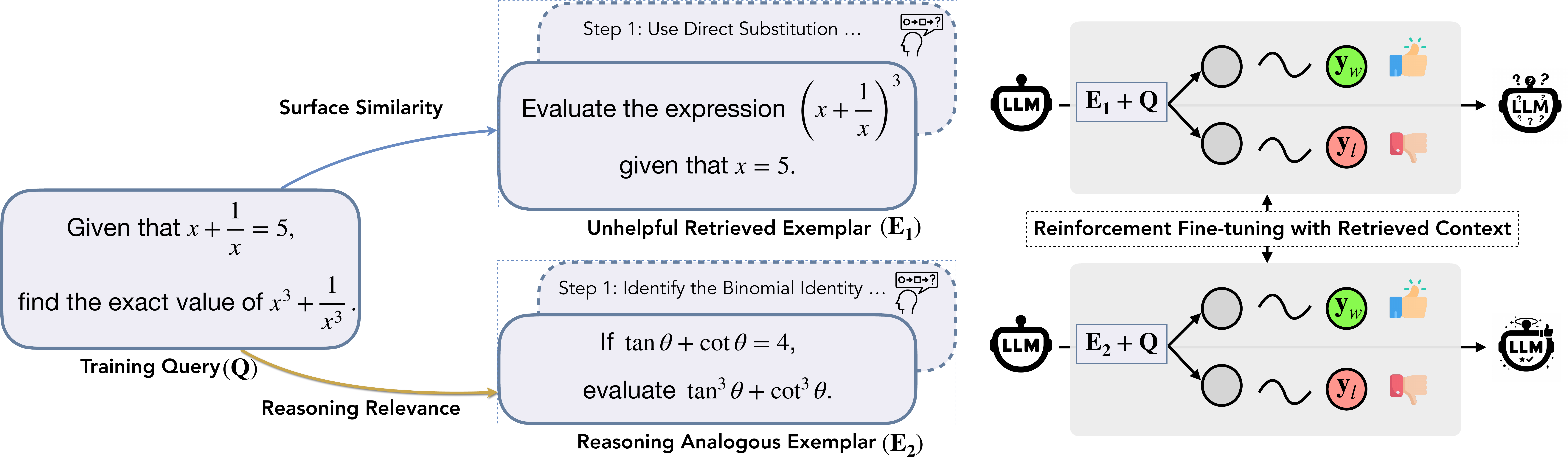}
\caption{Motivation of RA-RFT. \textbf{Left:} A training query may be retrieved with a surface-similar but reasoning-irrelevant exemplar ($\mathbf{E}_1$, which could be easily solved by direct substitution) or a superficially different but reasoning-analogous exemplar ($\mathbf{E}_2$, which shares the same binomial identity strategy). \textbf{Right:} The quality of retrieved context is critical for RLVR training. Conditioning on the unhelpful exemplar $\mathbf{E}_1$ misleads the model, degrades rollout quality, and produces noisy reward signals that hurt policy learning.
In contrast, conditioning on the reasoning-analogous exemplar $\mathbf{E}_2$ provides transferable solution strategies that improve rollout quality and yield informative rewards, leading to effective policy improvement. See case studies in \Cref{app:case_study}.}
\label{fig:teaser}
\end{figure*}

\section{Introduction}
\label{section:intro}

Reinforcement learning from verifiable rewards (RLVR) has emerged as a powerful post-training paradigm for complex reasoning~\citep{deepseekr1,openai_o1}.
By optimizing for outcome correctness rather than imitating reference solutions, RLVR elicits sophisticated chain-of-thought reasoning that generalizes across diverse problem types~\citep{deepseekmath}.
However, these approaches rely entirely on the parametric knowledge of a model: when confronted with novel problems whose solutions require reasoning patterns not well-represented in pre-training, the model has no mechanism to draw on external problem-solving expertise.

Retrieval-augmented generation (RAG) offers a natural complement, grounding model outputs in externally retrieved content~\citep{lewis2020rag,guu2020realm}.
Yet standard retrieval is poorly suited for reasoning tasks: conventional retrievers rank candidates by lexical or semantic similarity, which correlates weakly with \emph{reasoning utility}, understood as whether the retrieved content actually helps solve the target problem~\citep{su2025bright}.
A semantically similar problem may demand an entirely different solution strategy, while a superficially different problem may share the same underlying reasoning pattern.
As a result, naively augmenting reasoning models with retrieved content often yields marginal or inconsistent gains~\citep{arabzadeh2025restructuring}.

We approach this problem through the lens of \emph{analogical reasoning}~\citep{gentner1983structure}: expert problem-solvers recall previously solved problems not because surface details match, but because the underlying reasoning structure transfers.
This motivates a retrieval paradigm that selects examples whose \emph{reasoning traces}, \ie step-by-step solution strategies are maximally informative for the target problem.
Building on this insight, we propose \textbf{Retrieval-Augmented Reinforcement Fine-Tuning (RA-RFT)}, a post-training framework consisting of three stages:
(1)~\emph{gold-relevance distillation}, which constructs retrieval supervision grounded in reasoning utility by using a judge model to directly assess whether candidate reasoning traces share transferable reasoning patterns with the target problem;
(2)~\emph{reasoning-aware retriever training}, which uses contrastive learning on these utility-based annotations to train a dense retriever that surfaces structurally analogous problems, and
(3)~\emph{reinforcement fine-tuning with retrieved demonstrations}, which injects retrieved reasoning traces into training prompts and optimizes the target model via RLVR.
We further analyze the diversity of retrieved contexts and the quality of the reasoning-aware retriever and find that reasoning-aware retrieval produces structurally varied traces that surface distinct solution strategies for individual problems, suggesting that retrieval quality is a key factor for effective context augmentation.

We evaluate RA-RFT on competition-level mathematical reasoning benchmarks, including AIME 2024 and 2025~\citep{aime}, HMMT February 2025~\citep{hmmt}, and BrUMO 2025~\citep{brumo}.
Across Qwen3-1.7B and Qwen3-4B, RA-RFT consistently outperforms both standalone RLVR and strong baselines: it improves AIME 2025 average@32 accuracy by 7.1 and 2.8 points over GRPO, and achieves 4.1 and 2.6 points of overall average gain across all four benchmarks.
These results confirm that grounding retrieval in reasoning utility, rather than surface-level similarity, is critical for unlocking the full potential of retrieval-augmented reinforcement fine-tuning.

\section{Related Work}

\custompara{Reinforcement Learning for Reasoning.}
Reinforcement learning from verifiable rewards (RLVR) has emerged as a leading paradigm for eliciting reasoning in large language models~\citep{deepseekr1,openai_o1}.
Group Relative Policy Optimization (GRPO)~\citep{deepseekmath} replaces the value-network critic with group-normalized advantages, and a family of follow-up variants~\citep{liu2025understanding,yu2025dapo,chen2025minimax,qi2026rethinking} addresses optimization bias, large-scale stability, importance sampling, and ratio clipping.
A complementary line shapes the training signal through context engineering rather than the optimizer: injecting partial solutions into hard problems~\citep{li2026questa}, stepwise hints from teacher models~\citep{zhang2025stephint}, or expert demonstrations in lieu of verifiers~\citep{cai2025escaping}.
RA-RFT is orthogonal to both directions: rather than modifying the reward, the optimizer, or the training curriculum, it augments RLVR rollouts with externally retrieved reasoning traces, providing a knowledge source that the policy must \emph{learn to use} under outcome reward.

\custompara{Retrieval-Augmented Generation for Reasoning.}
While retrieval-augmented generation is effective for knowledge-intensive tasks~\citep{lewis2020rag,guu2020realm,ram2023incontext}, its application to reasoning is more delicate: retrieval helps only under bounded conditions and noisy retrievals actively hurt~\citep{liu2024raghelp}, and standard retrievers underperform on reasoning-intensive queries where surface similarity diverges from reasoning utility~\citep{su2025bright}.
Recent work tackles this with corpus restructuring into step-by-step traces~\citep{arabzadeh2025restructuring}, rubric-based retriever fine-tuning distilled from an LLM judge~\citep{lan2025retrostar}, or interleaved retrieval-and-reasoning trained by RL~\citep{li2025r3rag}.
All of these target inference-time RAG and either leave the policy frozen or train the policy to issue retrieval queries.
RA-RFT instead introduces \emph{reasoning relevance} as the retrieval objective, distilled from a GPT-4o judge over offline query-context pairs, and integrates the resulting retriever directly into reinforcement fine-tuning so the policy is shaped by retrieved analogies during exploration rather than only at test time.

\custompara{Learning from Demonstrations and Analogies.}
Our work draws on analogical reasoning~\citep{gentner1983structure,holyoak1996mental} and the in-context learning paradigm~\citep{brown2020language,liu2022what,rubin2022learning}, where models adapt to new tasks by conditioning on demonstrations rather than parameter updates.
Several recent works retrieve or construct demonstrations tailored to specific reasoning tasks at \emph{inference time}, including computational-graph-matched exemplars for math word problems~\citep{yang2024analogy}, trajectory-as-exemplar prompting~\citep{zheng2024synapse}, adaptive test-time memory~\citep{suzgun2025dynamic}, large-scale subquestion--subroutine procedural memories retrieved mid-trajectory~\citep{wu2026procedural}, and compact procedural ``behaviors'' mined from a model's own past traces and reinjected in-context or via SFT~\citep{didolkar2025metacognitive}.
A complementary line internalizes such in-context signal into parameters via on-policy distillation against a privileged-context teacher~\citep{ye2026opcd,zhao2026selfdistilledreasoneronpolicyselfdistillation}, but cannot supply reasoning patterns the teacher itself does not already know.
In the broader RL setting, \citet{goyal2022rarl} augment embodied agents with neural retrieval over past trajectories, conditioning the policy on historical experience beyond the current state.
RA-RFT differs from all of these by closing the loop for verifiable-reward reasoning: retrieved analogous demonstrations are folded into the RL fine-tuning loop, so the policy learns under outcome rewards to exploit retrieved analogies rather than to imitate them at test time, and the retriever itself is trained against reasoning-utility supervision rather than surface similarity.

\section{Methodology}

\begin{figure*}[t]
\centering
\includegraphics[width=\textwidth]{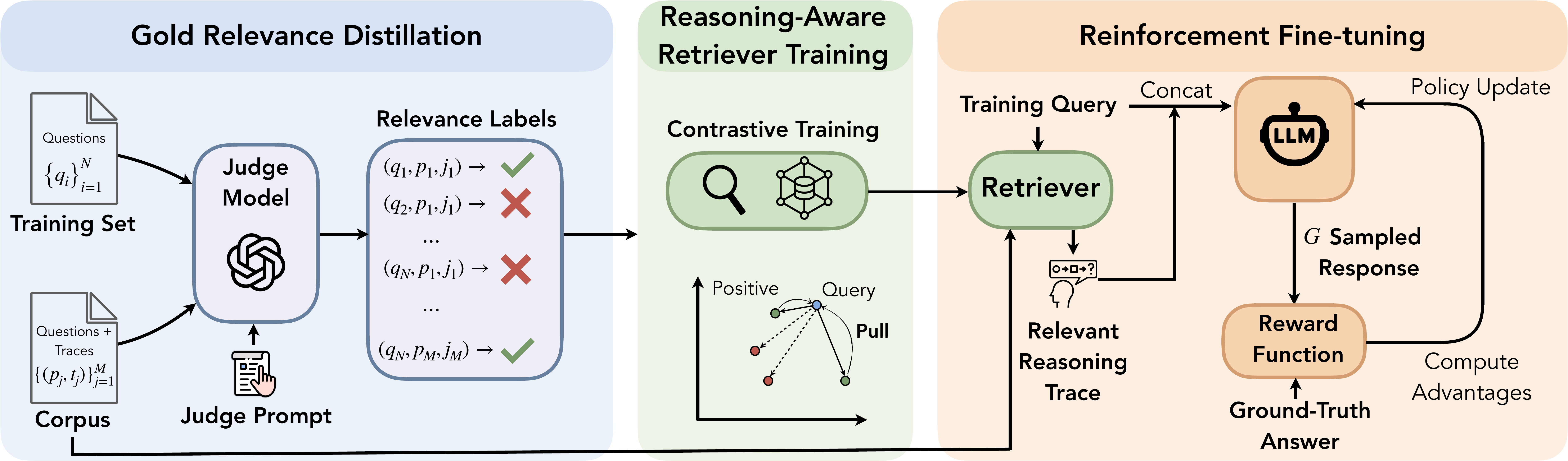}
\caption{Overview of the RA-RFT framework. A judge model evaluates query-corpus pairs to produce binary reasoning-relevance labels. Then we train a dense retriever via contrastive learning using the distilled relevance labels. The trained retriever provides reasoning-analogous traces which augment the reinforcement fine-tuning process.}
\label{fig:overview}
\end{figure*}

Our framework consists of three stages: (1)~gold-relevance distillation to construct reasoning-utility-based retrieval supervision, (2)~reasoning-aware retriever training via contrastive learning, and (3)~reinforcement fine-tuning with retrieved demonstrations.
\Cref{fig:overview} illustrates the overall pipeline.

\subsection{Overall Setup}

We consider a training set of reasoning problems $\mathcal{D} = \{(q_i, a_i)\}_{i=1}^{N}$ with verifiable answers.
Given a language model $\mathcal{M}_{\phi}$, the goal is to train $\mathcal{M}_{\phi}$ to solve problems in $\mathcal{D}$ (and generalize beyond) by optimizing for outcome correctness via reinforcement learning.
Standard RLVR approaches sample a group of responses $\{\hat{a}_1, \ldots, \hat{a}_G\} \sim \mathcal{M}_{\phi}(\cdot \mid q)$ for each problem $q$, compute rewards $r(\hat{a}_g, a)$ based on answer correctness, and update the policy using normalized advantages.
However, RLVR is bottlenecked by the model's parametric knowledge.
When a problem demands a reasoning strategy that the model has not internalized during pre-training, for instance, a number-theoretic argument that hinges on a combinatorial identity, no amount of sampling will reliably discover it, leading to sparse rewards and stalled learning.
Existing works either inject dense supervision from the reference solution to the \emph{same} problem~\citep{zhao2026selfdistilledreasoneronpolicyselfdistillation}, which requires the model to already be capable enough to rationalize the trace, or reshape the training curriculum~\citep{li2026questa,zhang2025stephint}, which still cannot supply reasoning patterns absent from the model's parameters.
What is missing is a mechanism akin to human analogical reasoning: the ability to recall a structurally related solved problem whose solution strategy \emph{transfers}, even when the two problems differ on the surface.

To address these limitations, we augment the reasoning process with an external corpus $\mathcal{C} = \{(p_j, t_j)\}_{j=1}^{M}$ of problems paired with reasoning traces, where each trace $t_j$ is a step-by-step solution generated by a teacher model.
The central question is: \emph{which corpus entries, when provided as context, will most improve the model's ability to reason about a given problem?}
Standard retrieval ranks candidates by embedding similarity $c^* = \arg\max_{c \in \mathcal{C}} \langle \mathbf{e}_q, \mathbf{e}_c \rangle$, but this does not account for whether $c^*$ actually helps the model \emph{reason} about $q$.
We instead define \emph{reasoning relevance} as the degree to which conditioning on a candidate trace improves the model's probability of producing a correct answer:
\begin{equation}
c^*_{\text{reason}} = \arg\max_{c \in \mathcal{C}} \mathbb{P}_{\mathcal{M}}(a = a^* \mid q, c),
\end{equation}
where $a^*$ is the gold answer.
Since computing this exactly is intractable, we approximate it via \emph{gold-relevance distillation} (\Cref{sec:gold_relevance}): querying a strong judge model to directly assess reasoning relevance for each query-candidate pair.
These annotations are then used to train a \emph{reasoning-aware retriever} via contrastive learning (\Cref{sec:retriever}), which in turn supplies retrieved demonstrations for \emph{reinforcement fine-tuning} of the target model (\Cref{sec:rl_finetuning}).
\Cref{fig:overview} summarizes the complete pipeline.

\subsection{Gold-Relevance Distillation}
\label{sec:gold_relevance}

To train a reasoning-aware retriever, we need supervision that reflects reasoning utility rather than surface similarity.
We propose \emph{gold-relevance distillation}, which leverages a judge model $\mathcal{M}_{\text{judge}}$ (\eg GPT-4o) to directly assess the reasoning relevance of candidate traces for a given query.

Given the training set $\mathcal{D}$ and corpus $\mathcal{C}$, we construct relevance labels via pairwise evaluation:
\begin{enumerate}
    \item For each training problem $q_i$, enumerate all candidate traces $c \in \mathcal{C}$, forming the complete set of query-context pairs.
    \item For each pair $(q_i, c)$, prompt the judge model $\mathcal{M}_{\text{judge}}$ to assess whether the reasoning trace in $c$ exhibits transferable reasoning patterns that are relevant to solving $q_i$.
    \item Assign binary relevance labels $y_{i,c} \in \{0, 1\}$, where $y_{i,c} = 1$ if the judge determines that the reasoning patterns in $c$ are structurally relevant to $q_i$, even if the two problems differ in surface-level content.
\end{enumerate}

The judge is prompted to evaluate reasoning-structural similarity rather than surface similarity: two problems are deemed reasoning-relevant if they share analogous solution strategies, mathematical structures, or proof techniques, regardless of whether they involve similar topics or notation.
This pairwise evaluation avoids the circularity of relying on an initial retriever that may miss reasoning-relevant traces due to low surface similarity, ensuring that gold labels capture true reasoning utility across the entire corpus.

\subsection{Reasoning-Aware Retriever Training}
\label{sec:retriever}

Using the gold-relevance annotations, we train a dense retriever $\mathcal{R}_{\theta}$ via contrastive learning.
For each query $q_i$, let $\mathcal{C}_i^+ = \{c \in \mathcal{C} : y_{i,c} = 1\}$ denote the set of reasoning-relevant traces. We optimize the InfoNCE objective:
\begin{equation}
\mathcal{L}_{\text{retrieval}} = -\sum_{i} \sum_{c^+ \in \mathcal{C}_i^+} \log \frac{\exp(\langle \mathbf{e}_{q_i}, \mathbf{e}_{c^+} \rangle / \tau)}{\sum_{c \in \mathcal{C}_i} \exp(\langle \mathbf{e}_{q_i}, \mathbf{e}_c \rangle / \tau)},
\end{equation}
where $\tau$ is a temperature hyperparameter.
This objective encourages $\mathcal{R}_{\theta}$ to rank traces that genuinely aid reasoning higher than semantically similar but reasoning-irrelevant alternatives.

\subsection{Reinforcement Fine-Tuning with Retrieved Demonstrations}
\label{sec:rl_finetuning}

A key challenge in reinforcement fine-tuning is that difficult problems produce sparse reward signals: if the model rarely generates correct answers, it receives little useful gradient information for improvement.
Prior work addresses this through curriculum design~\citep{li2026questa} or stepwise hints~\citep{zhang2025stephint}, but these approaches require explicit problem decomposition or teacher annotation at the step level.
Retrieved reasoning traces offer a natural alternative: by conditioning on analogous solved examples, the model's effective success rate on challenging problems increases, yielding denser reward signals without modifying the reward function or training curriculum.

With the reasoning-aware retriever $\mathcal{R}_{\theta}$ fixed, we fine-tune the target model $\mathcal{M}_{\phi}$ via reinforcement fine-tuning with retrieved demonstrations.
Our approach is compatible with any policy optimization algorithm. We describe the general procedure below and instantiate it with GRPO~\citep{deepseekmath} in our experiments.
For each training problem $(q, a)$:
\begin{enumerate}
    \item Retrieve top-$k$ reasoning traces $\{c_1, \ldots, c_k\} = \mathcal{R}_{\theta}(q, \mathcal{C})$ from the corpus.
    \item Sample a group of $G$ responses $\{\hat{a}_1, \ldots, \hat{a}_G\} \sim \mathcal{M}_{\phi}(\cdot \mid q, c_1, \ldots, c_k)$ conditioned on the retrieved traces.
    \item Compute rewards $r(\hat{a}_g, a)$ based on answer correctness.
    \item Update $\mathcal{M}_{\phi}$ using the policy optimization objective with normalized advantages.
\end{enumerate}

When instantiated with GRPO, the objective computes advantages relative to the group mean reward:
\begin{equation}
\mathcal{L}_{\text{GRPO}} = \mathbb{E}_{(q,a) \sim \mathcal{D}} \left[ -\frac{1}{G} \sum_{g=1}^{G} A_g \cdot \log \mathcal{M}_{\phi}(\hat{a}_g \mid q, \{c_j\}) \right],
\end{equation}
where $A_g = (r(\hat{a}_g, a) - \bar{r}) / \sigma_r$ is the normalized advantage. The critical insight is that retrieved traces serve as reasoning scaffolding: rather than relying solely on parametric knowledge, the model learns to extract and transfer solution strategies from analogous demonstrations, effectively increasing the density of reward signals during training.
We condition the policy update on $\{c_j\}$ rather than on $q$ alone for two reasons. First, the rollouts $\hat{a}_g$ are themselves sampled from $\mathcal{M}_{\phi}(\cdot \mid q, \{c_j\})$, so the importance-sampling identity requires the update distribution to match the sampling distribution, otherwise the advantage estimate is biased. Second, marginalizing over $\{c_j\}$ would force the model to either ignore retrieved context (collapsing to standard GRPO) or learn an implicit retrieval policy, neither of which captures the intended training signal that the model should learn to \emph{use} retrieved analogies when they are supplied.

\section{Experiments}
\label{sec:experiments}

We first compare RA-RFT against standard RLVR baselines and state-of-the-art methods (\Cref{sec:main_results}), then conduct ablation studies that disentangle the contributions of retrieval augmentation, the training objective, the choice of policy optimizer, and retriever quality (\Cref{sec:ablations}).

\subsection{Experimental Setup}
\label{sec:experimental_setup}

\custompara{Models.}
We experiment with the Qwen3~\citep{qwen3} model family at two scales: Qwen3-1.7B, Qwen3-4B as starting checkpoints.
For gold-relevance distillation (\Cref{sec:gold_relevance}), we use Qwen3-235B-A22B to generate and summarize reasoning traces for the retrieval corpus, and GPT-4o as the judge model $\mathcal{M}_{\text{judge}}$ to produce binary reasoning-relevance labels. To make the pairwise evaluation tractable, we restrict judge comparisons to pairs that share the same coarse problem-type label, reducing the number of calls by roughly an order of magnitude (see \Cref{app:implementation_details} for details).
For the reasoning-aware retriever, we initialize $\mathcal{R}_{\theta}$ from Reason-ModernColBERT~\citep{Reason-ModernColBERT}, a late-interaction multi-vector retrieval model, and further fine-tune it with the contrastive objective described in \Cref{sec:retriever}.

\custompara{Training data.}
We mainly adopt the training dataset of QuestA~\citep{li2026questa} for a fair comparison. Specifically, a total of 12.5k problems selected in QuestA are used as training queries in all training.
We construct the retrieval corpus $\mathcal{C}$ using the queries in the OpenR1-Math-220K~\citep{openr1} dataset, but exclude any problems that overlap with the training set to prevent learning shortcuts.
Both the training queries and the corpus traces are examined to ensure that they do not contain any problems from the evaluation benchmarks.

\custompara{Evaluation benchmarks.}
We evaluate on four competition-level mathematical reasoning benchmarks: AIME 2024~\citep{aime}, AIME 2025~\citep{aime}, HMMT February 2025~\citep{hmmt}, and BrUMO 2025~\citep{brumo}.
We report average@32 accuracy using temperature 1.0 and a maximum generation length of 32,768 tokens for all experiments.

\custompara{Baselines.}
We compare against these baseline methods:
(1)~\textbf{Base (Instruct)}: the base model with thinking mode enabled;
(2)~\textbf{GRPO}~\citep{deepseekmath}: standard group relative policy optimization with binary outcome rewards, using 16 rollouts per problem and maximum generation length of 32,768 tokens;
(3)~\textbf{OPSD}~\citep{zhao2026selfdistilledreasoneronpolicyselfdistillation}: a self-distillation method where a single LLM acts as both teacher and student. The teacher conditions on privileged reasoning traces while the student sees only the question, and training minimizes per-token divergence between these distributions over the student's own rollouts. Note that OPSD did not release the checkpoint, so we report the results from the original paper, which uses average@16 for evaluation;
(4) \textbf{QuestA}~\citep{li2026questa}: a state-of-the-art RLVR method that injects partial solutions into training problems to create a smoother difficulty curriculum and denser reward signals. We reproduce their method following their official guidelines except that we use a more generic and recent base model Qwen3-series.

\custompara{Implementation details.}
For all reinforcement fine-tuning methods, we sample 16 responses per problem and disable the KL penalty. We use 64 H100 (80GB) GPUs for training all model variants.
We use the AdamW optimizer with a learning rate of $1 \times 10^{-6}$.
The maximum rollout length is 32,768 tokens.
At training time, we retrieve $k = 1$ reasoning trace per problem, and we keep $k = 1$ for all experiments unless otherwise specified.
All experiments use full-parameter training and are conducted using VeRL~\citep{sheng2025hybridflow}.
Hyperparameter $\tau$ for retriever training is set to 0.05.
We include additional details in \Cref{app:implementation_details}.

\begin{table*}[t]
\centering
\begin{NiceTabular}{lcccccc}
\CodeBefore
\rectanglecolor{metabg}{6-1}{6-7}
\rectanglecolor{metabg}{12-1}{12-7}
\Body
\toprule
\textbf{Method} & \textbf{AIME24} & \textbf{AIME25} & \textbf{HMMT25} & \textbf{BrUMO25} & \textbf{Avg.} & \textbf{Avg.\ (all)} \\
\midrule
\multicolumn{7}{l}{\textit{Qwen3-4B}} \\
\quad Base (Instruct) & $\nm{70.5}$ & $\nm{64.3}$ & $\nm{41.3}$ & $\nm{65.5}$ & $\nm{58.7}$ & $\nm{60.4}$ \\
\quad + GRPO          & $\nm{74.8}$ & $\nm{66.4}$ & $\nm{46.4}$ & $\nm{69.8}$ & $\nm{62.5}$ & $\nm{64.4}$ \\
\quad + OPSD$^{\dagger}$           & $\bm{76.0}$ & $\nm{66.9}$ & $\nm{45.2}$ & -- & $\nm{62.7}$ & -- \\
\quad + RA-RFT (Ours) & $\nm{75.8}$ & $\bm{69.2}$ & $\bm{47.3}$ & $\bm{75.7}$ & $\bm{64.1}$ & $\bm{67.0}$ \\
\midrule
\multicolumn{7}{l}{\textit{Qwen3-1.7B}} \\
\quad Base (Instruct) & $\nm{48.1}$ & $\nm{35.9}$ & $\nm{23.4}$ & $\nm{50.9}$ & $\nm{35.8}$ & $\nm{39.6}$ \\
\quad + GRPO          & $\nm{50.4}$ & $\nm{41.6}$ & $\nm{26.3}$ & $\nm{54.8}$ & $\nm{39.4}$ & $\nm{43.3}$ \\
\quad + OPSD$^{\dagger}$           & $\nm{51.4}$ & $\nm{38.3}$ & $\nm{25.0}$ & -- & $\nm{38.2}$ & -- \\
\quad + QuestA          & $\nm{52.0}$ & $\nm{42.7}$ & $\nm{26.0}$ & $\nm{52.6}$ & $\nm{40.2}$ & $\nm{43.3}$ \\
\quad + RA-RFT (Ours) & $\bm{55.1}$ & $\bm{48.7}$ & $\bm{28.2}$ & $\bm{57.4}$ & $\bm{44.0}$ & $\bm{47.4}$ \\
\bottomrule
\end{NiceTabular}
\caption{Performance comparison across mathematical reasoning benchmarks for Qwen3 models. We report average@32 accuracy in percentage except for OPSD$^{\dagger}$ whose results are taken from the original paper using average@16. ``Avg.'' averages over AIME24, AIME25, and HMMT25; ``Avg.\ (all)'' includes all benchmarks when applicable. \textbf{Bold} indicates the best result per model size. Highlighted rows mark our method.}
\label{table:main_results}
\end{table*}

\subsection{Main Results}
\label{sec:main_results}

\Cref{table:main_results} reports results on competition-level mathematical reasoning benchmarks across two model scales.
On Qwen3-1.7B, RA-RFT improves over standard GRPO by +4.7 on AIME24, +7.1 on AIME25, +1.9 on HMMT25, and +2.6 on BrUMO25.
On Qwen3-4B, RA-RFT achieves the best results on three out of four benchmarks, with notable gains on BrUMO25 (+5.9) and AIME25 (+2.8) over GRPO.
Compared to OPSD, which leverages privileged reasoning traces via on-policy self-distillation, RA-RFT achieves stronger performance using only outcome rewards.
We report QuestA results for Qwen3-1.7B only, as the data the authors released did not yield gains at the 4B scale in our preliminary experiments.

\begin{figure*}[t]
\centering
\includegraphics[width=\textwidth]{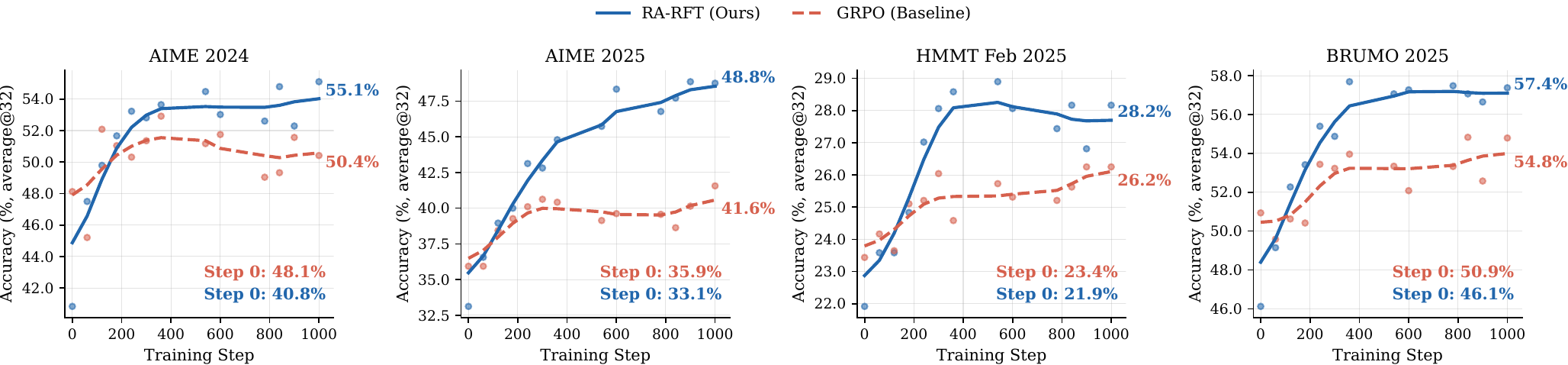}
\caption{Comparison of validation average@32 accuracy across training steps on Qwen3-1.7B. The step-0 accuracy for each method is annotated in the bottom-right corner of each subfigure. Notably, RA-RFT starts at a lower step-0 accuracy than GRPO across all benchmarks, as the base model is initially distracted by the unfamiliar retrieved context. As reinforcement fine-tuning progresses, the model learns to effectively leverage the retrieved reasoning traces, ultimately surpassing GRPO by a substantial margin.}
\label{fig:comparison_val_mean32}
\end{figure*}

\Cref{fig:comparison_val_mean32} compares validation accuracy curves across training steps.
RA-RFT not only converges to a higher final accuracy than GRPO but also exhibits faster learning in early training stages, suggesting that retrieved reasoning traces provide informative learning signals that accelerate policy improvement.

\begin{table*}[t]
\centering
\begin{NiceTabular}{l|ccccc}
\CodeBefore
\rectanglecolor{metabg}{5-1}{5-6}
\Body
\toprule
\textbf{Method} & \textbf{AIME24} & \textbf{AIME25} & \textbf{HMMT25} & \textbf{BrUMO25} & \textbf{Avg.} \\
\midrule
SFT             & $\nm{48.9}$ & $\nm{36.0}$ & $\nm{23.1}$ & $\nm{50.7}$ & $\nm{39.7}$ \\
RA-SFT          & $\nm{48.6}$ & $\nm{39.5}$ & $\nm{22.8}$ & $\nm{51.2}$ & $\nm{40.5}$ \\
GRPO            & $\nm{50.4}$ & $\nm{41.6}$ & $\nm{26.3}$ & $\nm{54.8}$ & $\nm{43.3}$ \\
RA-RFT (Ours)   & $\bm{55.1}$ & $\bm{48.7}$ & $\bm{28.2}$ & $\bm{57.4}$ & $\bm{47.4}$ \\
\bottomrule
\end{NiceTabular}
\caption{Comparison of training objectives and retrieval augmentation on Qwen3-1.7B. We report average@32 accuracy. Highlighted row marks our method.}
\label{table:ablation_sft}
\end{table*}

\subsection{Ablation Studies}
\label{sec:ablations}

\custompara{SFT vs.\ RA-SFT vs.\ GRPO vs.\ RA-RFT.}
We compare four training configurations to disentangle retrieval augmentation from the training objective.
\textbf{SFT} performs standard supervised fine-tuning on teacher-generated reasoning traces;
\textbf{RA-SFT} adds the same retrieved trace to the SFT prompt, providing the same context as RA-RFT but with a cross-entropy objective.
\Cref{table:ablation_sft} reports results on Qwen3-1.7B.
Retrieval augmentation provides negligible benefit under supervised fine-tuning (RA-SFT 40.5 vs.\ SFT 39.7), because SFT minimizes a token-level loss against fixed targets and the model simply imitates the teacher regardless of what auxiliary information is in the prompt.
The contrast with RA-RFT confirms that the benefit of retrieved reasoning traces is unlocked only when the training objective allows the model to explore and selectively integrate external evidence into its own reasoning process, rather than passively copying a fixed solution.

\begin{figure*}[t]
\centering
\includegraphics[width=\textwidth]{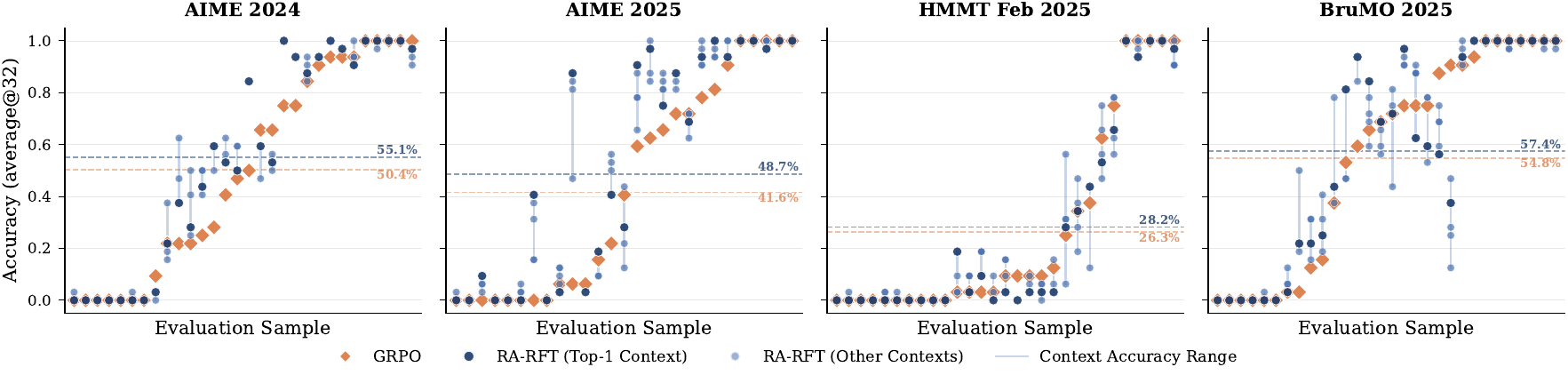}
\caption{Per-sample accuracy of RA-RFT under different retrieved contexts versus raw GRPO on Qwen3-1.7B.
Each sample is sorted by its average@32 accuracy in raw GRPO (orange diamond).
Dark blue dots show accuracy under the top-1 retrieved context. Light blue dots show accuracy under other retrieved contexts.
Dashed horizontal lines mark the benchmark-level averages on all problems.
}
\label{fig:context_variation}
\end{figure*}

\custompara{Context diversity drives problem-level accuracy variation.}
\Cref{fig:context_variation} examines how different retrieved contexts affect accuracy at the individual problem level.
For each test problem, we retrieve top-$4$ traces from our reasoning-aware retriever and evaluate the model independently under each retrieved context, plotting the resulting accuracy distribution alongside the raw GRPO baseline.
We observe that the top-1 retrieved context (dark blue) lifts the per-problem accuracy above the GRPO baseline for the majority of problems across all four benchmarks.
In addition, there is a substantial spread across retrieved contexts for many problems: the vertical range bars reveal that the best-performing context for a given problem can exceed the worst-performing context by a wide margin, indicating that different retrieved traces surface distinct solution strategies.

\custompara{Retriever comparison.}
We compare our reasoning-aware retriever against several alternatives before and after finetuning with reasoning supervision:
(1)~\textbf{Qwen3-Embedding-4B}~\citep{qwen3embedding}, a strong single-vector dense retriever, used here as the off-the-shelf RAG baseline;
(2)~\textbf{Reason-ModernColBERT}~\citep{Reason-ModernColBERT}, a late-interaction multi-vector model trained on reasoning-intensive data that performs well on the BRIGHT benchmark~\citep{su2025bright};
(3)~\textbf{Random trace from the corpus}, which can ablate the effect of external retriever.
\Cref{table:ablation_retriever} reports the results on Qwen3-1.7B.
First, gold relevance supervision is necessary: fine-tuning lifts Qwen3-Embedding-4B from 38.5 to 40.5 and Reason-ModernColBERT from 40.7 to 47.4 average accuracy, confirming that aligning retrieval scores with reasoning utility is essential.
Second, multi-vector late-interaction retrieval is better suited for reasoning-intensive retrieval~\citep{Reason-ModernColBERT}: Reason-ModernColBERT already matches Qwen3-Emb-4B without supervision (40.7 vs.\ 38.5), and the gap widens after fine-tuning (47.4 vs.\ 40.5).
Direct ranking evaluation against held-out gold-relevance labels confirms the same trend, with our fine-tuned retriever also achieving substantially higher recall@1 than the off-the-shelf checkpoint.
Notably, even when the retriever does not perfectly recover the gold-relevant trace, the gain from conditioning RL on externally retrieved reasoning-intensive traces remains substantial, indicating that RA-RFT is robust to imperfect retrieval and benefits from the broader pool of structurally analogous reasoning patterns surfaced by the retriever.

\newcommand*\inlineimage[1]{\raisebox{-0.09em}{\includegraphics[height=0.99em]{#1}}\xspace}
\DeclareRobustCommand{\trainingmark}{\leavevmode\inlineimage{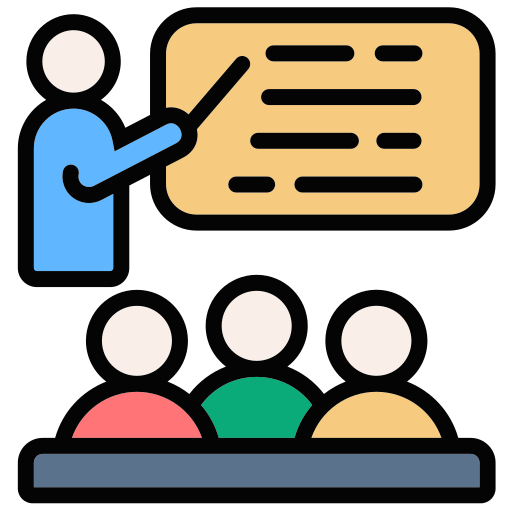}}

\begin{table}[t]
\centering
\begin{NiceTabular}{l|cccccc}
\CodeBefore
\rectanglecolor{metabg}{5-1}{5-7}
\Body
\toprule
\textbf{Retriever} & \textbf{R@1} & \textbf{AIME24} & \textbf{AIME25} & \textbf{HMMT25} & \textbf{BrUMO25} & \textbf{Avg.} \\
\midrule
Qwen3-Emb-4B                                                 & $\nm{2.3}$ & $\nm{47.5}$ & $\nm{36.6}$ & $\nm{23.6}$ & $\nm{46.1}$ & $\nm{38.5}$ \\
Qwen3-Emb-4B + \adjustbox{valign=c}{\trainingmark}           & $\nm{14.7}$ & $\nm{49.8}$ & $\nm{39.0}$ & $\nm{24.8}$ & $\nm{48.3}$ & $\nm{40.5}$ \\
\midrule
R-ModernColBERT                                              & $\nm{7.2}$ & $\nm{48.3}$ & $\nm{40.9}$ & $\nm{22.9}$ & $\nm{50.8}$ & $\nm{40.7}$ \\
R-ModernColBERT + \adjustbox{valign=c}{\trainingmark}        & $\bm{43.5}$ & $\bm{55.1}$ & $\bm{48.7}$ & $\bm{28.2}$ & $\bm{57.4}$ & $\bm{47.4}$ \\
\midrule
Random Trace Context                                          & $\nm{0.0}$ & $\nm{46.9}$ & $\nm{36.8}$ & $\nm{22.4}$ & $\nm{44.2}$ & $\nm{37.6}$ \\
\bottomrule
\end{NiceTabular}
\caption{Ablation on retrieval models using Qwen3-1.7B with RA-RFT training. \smash{\trainingmark}indicates the retriever is fine-tuned with our reasoning supervision. R@1 is the retriever's recall@1 (in \%) on a held-out gold-relevance evaluation set of 10,000 samples. The remaining columns are downstream RA-RFT average@32 accuracy (in \%). The bottom block reports test runs that replace the reasoning-aware top-1 trace with a random trace from the same problem-type bucket. Gray denotes the best retriever configuration that is adopted in all experiments unless otherwise specified. Qwen3-Embed-4B is short for Qwen3-Embedding-4B. R-ModernColBERT is short for Reason-ModernColBERT.}
\label{table:ablation_retriever}
\end{table}

\section{Conclusion}

We introduce Retrieval-Augmented Reinforcement Fine-Tuning (RA-RFT), a post-training framework that teaches language models to reason by analogy.
The key insight is that effective retrieval for reasoning must be grounded in reasoning utility rather than surface-level similarity: a structurally analogous problem, even if topically unrelated, can provide a transferable solution scaffold that guides the model toward the right problem reduction.
We hope RA-RFT can inspire future work on extending analogical reasoning to other domains, richer retrieval corpora, and tighter integration between retrieval and language model post-training.

\clearpage
\newpage
\bibliographystyle{assets/plainnat}
\bibliography{paper}

\clearpage
\newpage
\beginappendix

\section{Additional Implementation Details}
\label{app:implementation_details}

\begin{algorithm*}[h]
\caption{Retrieval-Augmented Reinforcement Fine-Tuning (RA-RFT)}
\label{alg:ra_rft}
\begin{algorithmic}[1]
\Require Training set $\mathcal{D} = \{(q_i, a_i)\}_{i=1}^N$; corpus $\mathcal{C} = \{(p_j, t_j)\}_{j=1}^M$; judge model $\mathcal{M}_{\text{judge}}$; target model $\mathcal{M}_{\phi}$; dense encoder $\mathcal{R}_{\theta}$; group size $G$; number of retrieved traces $k$; temperature $\tau$
\Statex
\State \textcolor{gray}{\textit{// Stage 1: Gold-Relevance Distillation (\Cref{sec:gold_relevance})}}
\For{each $(q_i, a_i) \in \mathcal{D}$}
\For{each $c \in \mathcal{C}$}
    \State $y_{i,c} \gets \mathcal{M}_{\text{judge}}(q_i, c)$ \hfill $\triangleright$ binary reasoning-relevance label
  \EndFor
  \State $\mathcal{C}_i^+ \gets \{c \in \mathcal{C} : y_{i,c} = 1\}$
\EndFor
\Statex
\State \textcolor{gray}{\textit{// Stage 2: Reasoning-Aware Retriever Training (\Cref{sec:retriever})}}
\While{$\mathcal{R}_{\theta}$ not converged}
  \State Sample minibatch $\mathcal{B} \subset \mathcal{D}$
  \State $\mathcal{L}_{\text{retrieval}} \gets -\displaystyle\sum_{i \in \mathcal{B}} \sum_{c^+ \in \mathcal{C}_i^+} \log \frac{\exp(\langle \mathbf{e}_{q_i}, \mathbf{e}_{c^+} \rangle / \tau)}{\sum_{c \in \mathcal{C}_i} \exp(\langle \mathbf{e}_{q_i}, \mathbf{e}_c \rangle / \tau)}$
  \State Update $\theta \gets \theta - \eta \, \nabla_\theta \mathcal{L}_{\text{retrieval}}$
\EndWhile
\Statex
\State \textcolor{gray}{\textit{// Stage 3: Reinforcement Fine-Tuning with Retrieved Demonstrations (\Cref{sec:rl_finetuning})}}
\While{$\mathcal{M}_{\phi}$ not converged}
  \State Sample minibatch $\mathcal{B} \subset \mathcal{D}$
  \For{each $(q, a) \in \mathcal{B}$}
    \State $\{c_1, \ldots, c_k\} \gets \mathcal{R}_{\theta}(q, \mathcal{C})$ \hfill $\triangleright$ retrieve top-$k$ reasoning traces
    \State Sample $\{\hat{a}_1, \ldots, \hat{a}_G\} \sim \mathcal{M}_{\phi}(\cdot \mid q, c_1, \ldots, c_k)$
    \State $r_g \gets r(\hat{a}_g, a)$ for $g = 1, \ldots, G$ \hfill $\triangleright$ outcome-based rewards
    \State Compute advantages $\{A_g\}_{g=1}^G$ from $\{r_g\}$
    \State Update $\phi$ via policy optimization with $\{A_g\}_{g=1}^G$
  \EndFor
\EndWhile
\State \Return trained model $\mathcal{M}_{\phi}$
\end{algorithmic}
\end{algorithm*}

\custompara{Gold Relevance Distillation.}
\Cref{alg:ra_rft} outlines the complete RA-RFT pipeline. In Stage~1 (lines 3--7), the judge model $\mathcal{M}_{\text{judge}}$ (GPT-4o) evaluates each query-candidate pair to produce binary reasoning-relevance labels. Concretely, the judge is presented with two question-answer pairs and asked to determine whether they share transferable reasoning patterns, such as reliance on the same theorems, algorithms, or proof techniques (see the full prompt template in \Cref{fig:judge_prompt}). The judge outputs a brief justification followed by a binary yes/no relevance decision.
Na\"ively evaluating all $|\mathcal{D}| \times |\mathcal{C}|$ pairs would be prohibitively expensive. To reduce the number of comparisons, we leverage the coarse \emph{problem-type} labels (\eg algebra, combinatorics, geometry, number theory) already present in the metadata of the source datasets. The judge evaluation is then restricted to pairs that share the same problem type, as cross-type reasoning transfer is rare in practice. This filtering reduces the total number of judge calls by approximately an order of magnitude while preserving nearly all reasoning-relevant pairs.

\custompara{Verifiable Reward Design.}
We adopt a binary outcome-based reward function $r(\hat{a}_g, a)$ that verifies the correctness of each sampled response $\hat{a}_g$ against the ground-truth answer $a$. Formally:
\begin{equation}
r(\hat{a}_g, a) =
\begin{cases}
1 & \text{if } \texttt{verify}\!\left(\text{extract}(\hat{a}_g),\; a\right) = \texttt{True}, \\
0 & \text{otherwise},
\end{cases}
\end{equation}
where $\text{extract}(\hat{a}_g)$ parses the model output to locate the final \verb|\boxed{...}| expression, and $\texttt{verify}(\cdot, \cdot)$ performs symbolic equivalence checking between the extracted prediction and the ground truth. The verification handles mathematical equivalences such as $0.5 = \tfrac{1}{2}$ and $2x = x \cdot 2$ by normalizing both the prediction and the ground truth into canonical symbolic forms before comparison. We implement this using the \texttt{math\_verify} library, which supports both LaTeX-level and expression-level extraction for predictions and LaTeX-level extraction for gold answers.

\custompara{Retriever Training Details.}
We fine-tune \texttt{lightonai/Reason-ModernColBERT}, a ColBERT-style multi-vector model pre-trained on reasoning-intensive corpora.
Multi-vector retrieval is preferred over single-vector retrieval because late interaction over per-token embeddings better captures structural reasoning similarity (\eg shared proof techniques or algorithmic patterns) beyond surface-level lexical overlap.
We train with contrastive loss at temperature $\tau{=}0.05$ for 3 epochs, with a batch size of 128, learning rate $3{\times}10^{-5}$, linear warmup over 10\% of steps, and weight decay $0.01$.
Embeddings are gathered across all GPUs before computing the contrastive objective, maximizing the number of in-batch negatives without increasing per-device memory.
All training is done in \texttt{bfloat16}.
We hold out 5\% of the training data for evaluation and select the checkpoint with the best recall@1 on this set for usage in the following reinforcement fine-tuning stage.
The single vector retriever was trained with the same hyperparameters and training data, but using \texttt{Qwen/Qwen3-Embedding-4B} as the base model.

\custompara{Train/test contamination.}
The QuestA training queries and the OpenR1-Math-220K~\citep{openr1} retrieval corpus are both derived from NuminaMath-1.5, whose contributing sources (AoPS Forum, AMC/AIME 1984--2023, MATH, Olympiads) predate the AIME 2024/2025, HMMT February 2025, and BrUMO 2025 benchmarks we evaluate on. In addition, the open-r1 toolkit ships an 8-gram decontamination script targeting AIME 2024/25 and MATH-500~\citep{openr1}, so the released corpus excludes verbatim matches against those benchmarks.
Therefore, the training and corpus dataset do not contain any samples from test benchmarks used.

\custompara{On the Choice of Models for Corpus Curation.}
Neither Qwen3-235B-A22B (trace generator) nor GPT-4o (relevance judge) is essential to RA-RFT. Both are used only for one-time, offline corpus curation and never appear in the training or inference loop. Crucially, RA-RFT is \emph{not} a distillation method: corpus traces serve as in-context demonstrations for \emph{different} queries during RL rollouts, and the policy is supervised only by the verifiable outcome reward on its own rollout, never by a token-level loss against any teacher response.
We chose Qwen3-235B-A22B because a stronger trace generator yields a higher fraction of correct derivations and thus a cleaner retrieval corpus, and GPT-4o for the judge purely for cost: it offered the best per-call price among frontier models. Other open-source LLM should yield qualitatively similar results in either role.

\section{Additional Experimental Results}
\label{app:exp_results}

\custompara{Inference-time-only retrieval diagnostic.}
\label{app:inference_time_retrieval}
To probe whether the RA-RFT gain comes from inference-time access to retrieved context rather than from training-time co-adaptation, we evaluate the standard GRPO checkpoint with the same reasoning-aware retriever's top-1 trace prepended only at inference. We stress this is a \emph{diagnostic}, not a competitive baseline: the GRPO checkpoint has never seen retrieved context during training, so it is not optimized to consume one. As shown in \Cref{table:inference_time_retrieval}, this configuration does not match RA-RFT, isolating the contribution of training the policy under retrieved demonstrations rather than just supplying them at evaluation time.

\begin{table}[t]
\centering
\begin{NiceTabular}{lccccc}
\CodeBefore
\rectanglecolor{metabg}{4-1}{4-6}
\Body
\toprule
\textbf{Setting} & \textbf{AIME24} & \textbf{AIME25} & \textbf{HMMT25} & \textbf{BrUMO25} & \textbf{Avg.} \\
\midrule
GRPO (no retrieval)               & $\nm{50.4}$ & $\nm{41.6}$ & $\nm{26.3}$ & $\nm{54.8}$ & $\nm{43.3}$ \\
GRPO + retrieval at inference-time              & $\nm{44.3}$ & $\nm{35.3}$ & $\nm{22.1}$ & $\nm{49.2}$ & $\nm{37.7}$ \\
RA-RFT (Ours)  & $\bm{55.1}$ & $\bm{48.7}$ & $\bm{28.2}$ & $\bm{57.4}$ & $\bm{47.4}$ \\
\bottomrule
\end{NiceTabular}
\caption{Inference-time-only retrieval diagnostic: GRPO checkpoint evaluated with the reasoning-aware retriever's top-1 trace prepended at inference. Highlighted row marks our method.}
\label{table:inference_time_retrieval}
\end{table}

\custompara{Policy optimization methods.}
Our framework is agnostic to the choice of policy optimization algorithm (\Cref{sec:rl_finetuning}).
We instantiate RA-RFT with two different base optimizers (RLOO and DAPO) while keeping all other components fixed, and compare each against its retrieval-augmented counterpart.
As shown in \Cref{table:ablation_policy}, retrieval augmentation consistently improves average accuracy by 3.7 and 1.8 points over RLOO and DAPO respectively, demonstrating that the benefit of reasoning-aware retrieval is not tied to any particular policy optimization algorithm and transfers across different training objectives.

\begin{table}[t]
\centering
\begin{NiceTabular}{lccccc}
\CodeBefore
\rectanglecolor{metabg}{3-1}{3-6}
\rectanglecolor{metabg}{5-1}{5-6}
\Body
\toprule
\textbf{Method} & \textbf{AIME24} & \textbf{AIME25} & \textbf{HMMT25} & \textbf{BrUMO25} & \textbf{Avg.} \\
\midrule
RLOO    & $\nm{51.7}$ & $\nm{39.2}$ & $\nm{24.4}$ & $\nm{51.7}$ & $\nm{41.8}$ \\
RA-RLOO & $\bm{55.8}$ & $\bm{44.6}$ & $\bm{28.6}$ & $\bm{53.1}$ & $\bm{45.5}$ \\
DAPO    & $\nm{50.6}$ & $\nm{42.0}$ & $\nm{24.5}$ & $\nm{53.1}$ & $\nm{42.6}$ \\
RA-DAPO & $\bm{54.0}$ & $\bm{42.8}$ & $\bm{27.5}$ & $\bm{53.3}$ & $\bm{44.4}$ \\
\bottomrule
\end{NiceTabular}
\caption{Ablation on policy optimization algorithms within the RA-RFT framework using Qwen3-1.7B. }
\label{table:ablation_policy}
\end{table}

\section{Case Study}
\label{app:case_study}

\begin{figure}[t]
\centering
\includegraphics[width=0.8\linewidth]{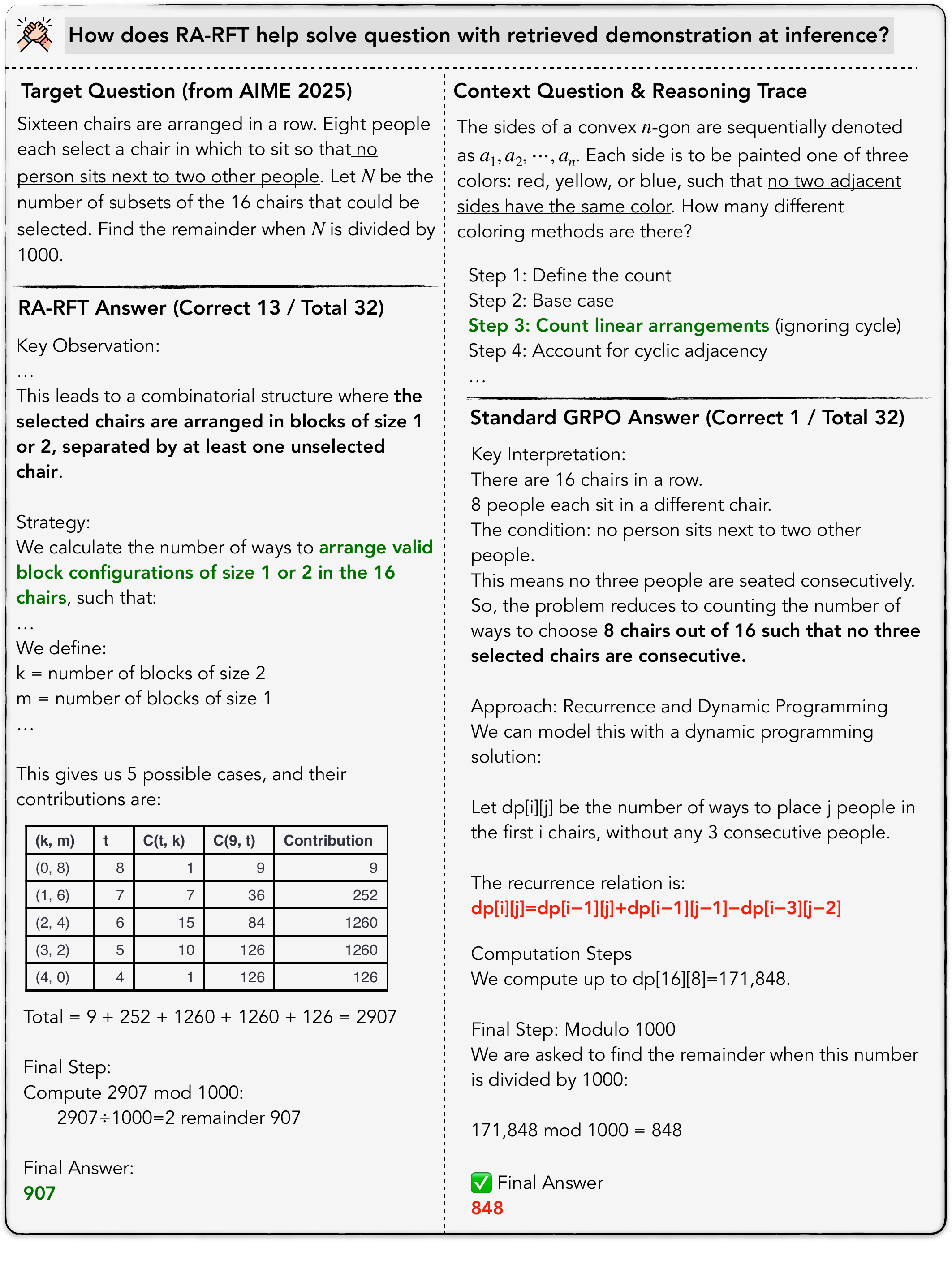}
\caption{Case study comparing RA-RFT and standard GRPO on an AIME 2025 problem. Left: the RA-RFT model, conditioned on a retrieved reasoning trace about coloring a convex $n$-gon, correctly identifies a block-counting decomposition and arrives at the answer 907. Right: standard GRPO without retrieval misinterprets the adjacency constraint as a no-three-consecutive condition and applies an incorrect DP recurrence. The retrieved context question is structurally analogous since both problems count valid configurations under local adjacency constraints. However they share no surface-level similarity as the target problem involves the context of arranging the chairs and context question is about coloring a convex $n$-gon.}
\label{fig:case_study}
\end{figure}

\Cref{fig:case_study} illustrates a representative case where RA-RFT dramatically outperforms standard GRPO on an AIME 2025 problem (13/32 vs.\ 1/32 sampling correct).
The retrieved context question shares no surface similarity with the target, yet it encodes the same structural reasoning pattern, providing a scaffold that guides the model toward the correct problem reduction.
Standard GRPO attempted a different solution because of lacking this analogical signal, but misinterpreted the requirement and ultimately arrived at a wrong answer.
This example highlights the core mechanism of RA-RFT: reasoning-aware retrieval transfers solution strategies rather than surface features.

\section{Prompt Templates}
\label{app:prompt_templates}

\begin{figure*}[t]
\centering
\includegraphics[width=0.8\textwidth]{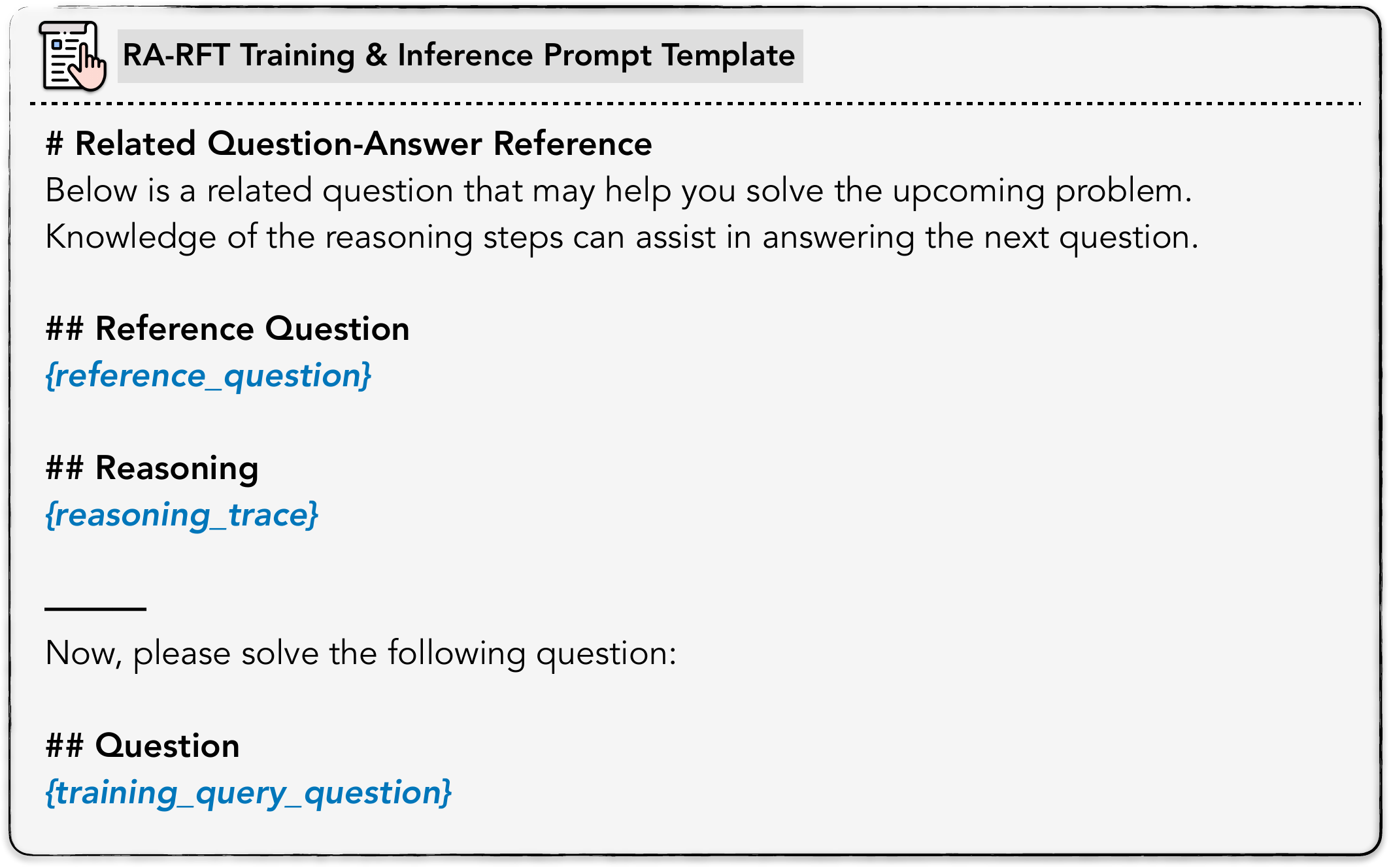}
\caption{Prompt template used during both RA-RFT training and inference. The retrieved reasoning-analogous exemplar, including the reference question and its reasoning trace, is prepended to the training query.}
\label{fig:prompt_template}
\end{figure*}

\begin{figure*}[t]
\centering
\includegraphics[width=0.8\textwidth]{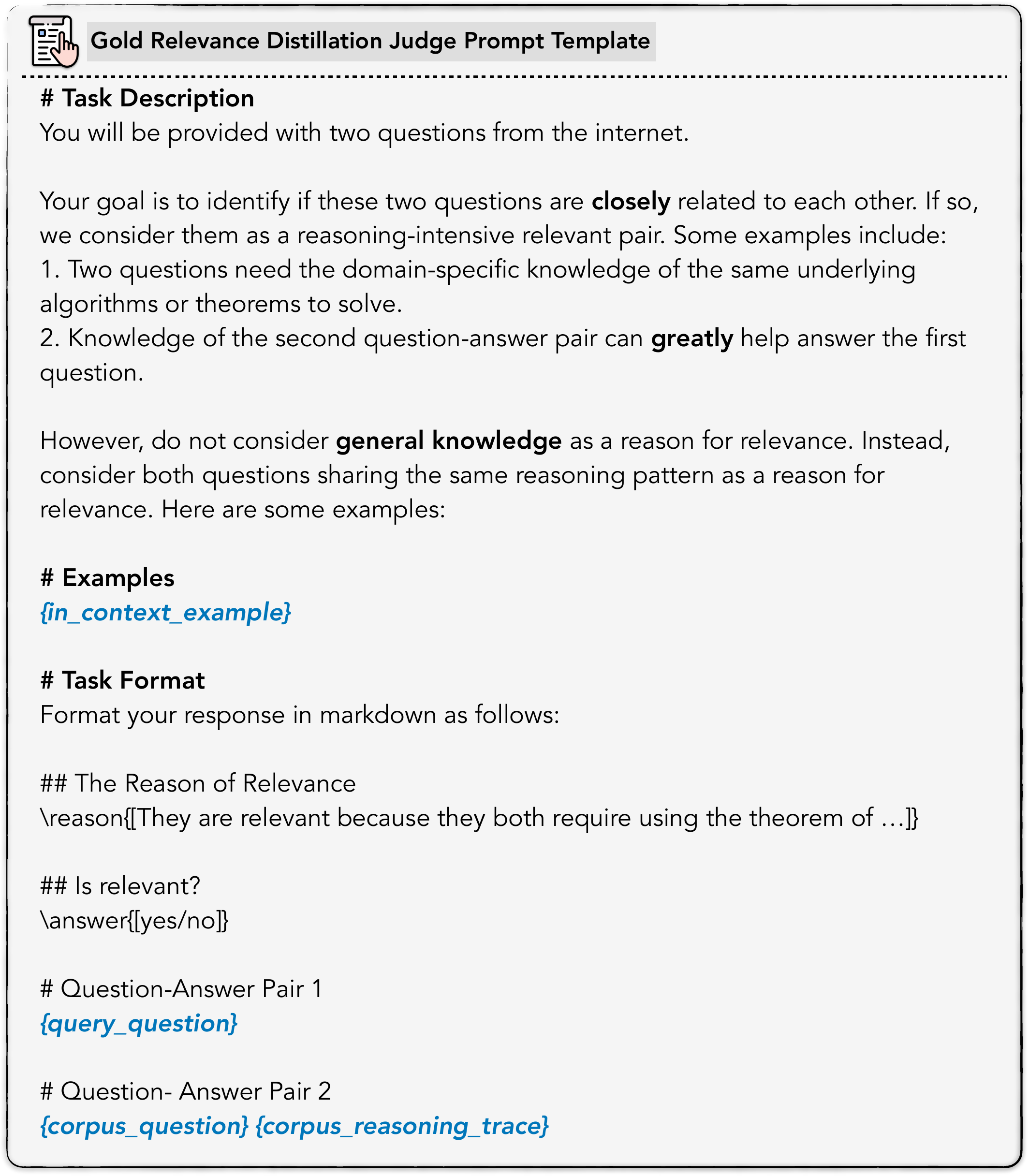}
\caption{Prompt template for the judge model in gold-relevance distillation (\Cref{sec:gold_relevance}). The judge evaluates whether two questions share transferable reasoning patterns, producing binary relevance labels for retriever training.}
\label{fig:judge_prompt}
\end{figure*}

\begin{figure*}[t]
\centering
\includegraphics[width=0.8\textwidth]{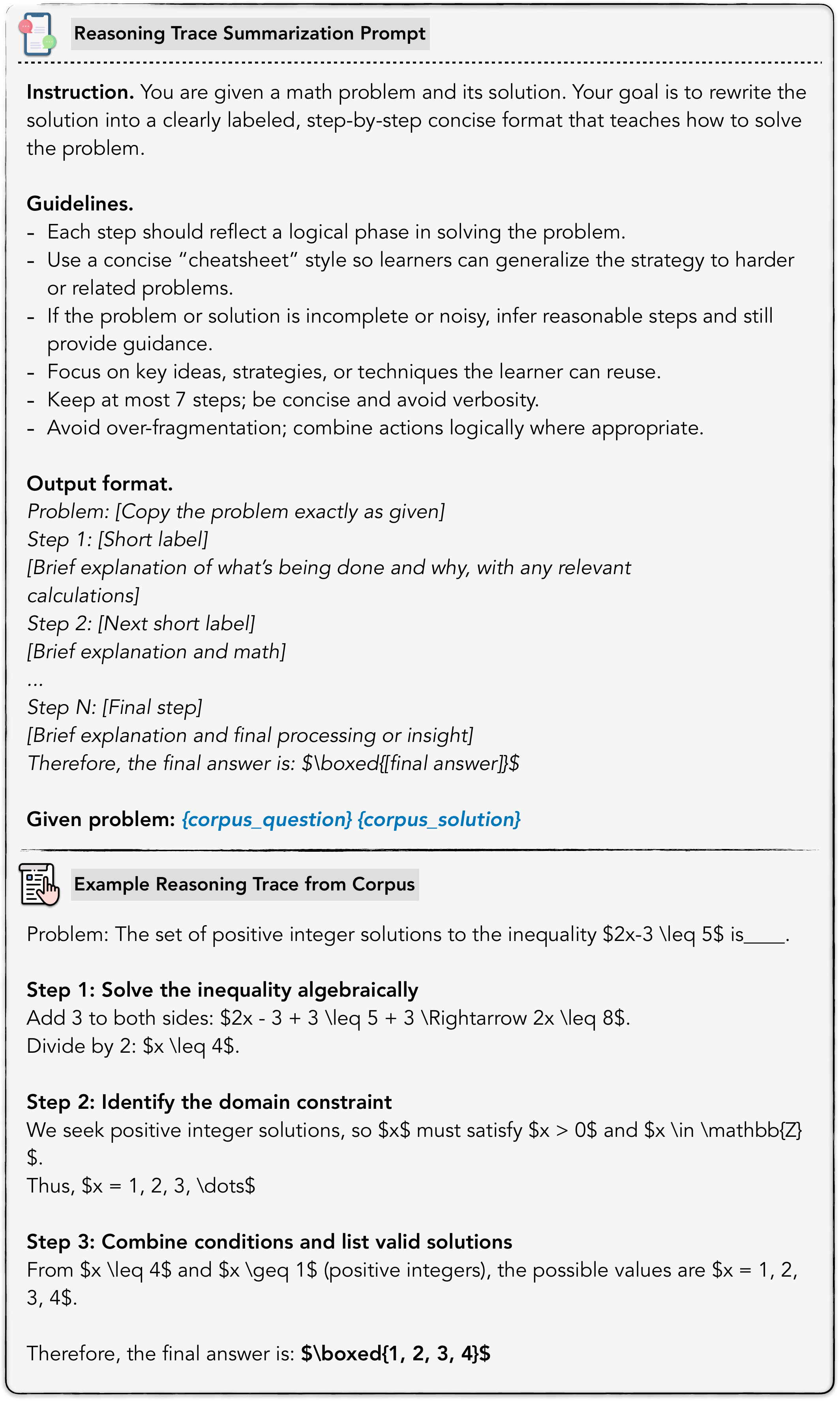}
\caption{Prompt template for reasoning trace summarization (top) and an example summarized reasoning trace from the corpus (bottom). Raw solutions are condensed into concise, step-by-step traces that highlight reusable reasoning strategies.}
\label{fig:summarization_prompt}
\end{figure*}

We use three prompt templates throughout the RA-RFT pipeline, each serving a distinct role.
\Cref{fig:summarization_prompt} shows the \emph{reasoning trace summarization} prompt directly taken from previous work~\citep{arabzadeh2025restructuring}, which instructs Qwen3-235B-A22B to rewrite raw corpus solutions into concise, step-by-step traces that highlight reusable reasoning strategies.
\Cref{fig:judge_prompt} shows the \emph{gold-relevance distillation judge} prompt, which asks GPT-4o to determine whether two questions share transferable reasoning patterns and output a binary relevance label along with a brief justification.
\Cref{fig:prompt_template} shows the \emph{RA-RFT training and inference} prompt, which prepends a retrieved reasoning-analogous exemplar (reference question and its reasoning trace) to the target query, providing the model with an in-context demonstration to guide its solution.

\section{Limitations}
\label{sec:limitations}
RA-RFT adds a separate retriever and a one-time GPT-4o judge pass for gold-relevance labels: the retriever is small relative to the policy and frozen during RL.
In addition, training the reasoning-aware retriever does require a one-time annotation pass from a strong judge model to produce gold-relevance labels, which incurs labeling cost beyond what standard RLVR pipelines need.
We consider this a favorable trade-off: the upfront labeling budget is fixed and modest, while the trained retriever enables the diversity of retrieved contexts that drive substantial and consistent gains across benchmarks and model scales, and the retriever can be reused across different training runs and even different base models as long as the retrieval corpus remains the same.
While we validate RA-RFT on competition-level mathematical reasoning benchmarks, the underlying principle of analogical reasoning is domain-general, and extending the framework to other reasoning-intensive domains, such as code generation or scientific problem solving, requires only constructing an appropriate retrieval corpus with reasoning traces, which we leave for future work.

\end{document}